\documentclass[11pt,a4paper]{article}
\usepackage[hyperref]{acl2020}
\usepackage{times}
\usepackage{latexsym}

\usepackage{microtype}

\aclfinalcopy %
\def\aclpaperid{1519} %

\usepackage{graphicx}
\usepackage{tabularx}
\usepackage{soul}

\usepackage{epstopdf}
\usepackage[utf8]{inputenc}

\usepackage{xstring}
\usepackage{xcolor}
\usepackage{booktabs}
\usepackage{xspace}

\newcommand{\asset}{ASSET\xspace}
\newcommand{\turk}{TurkCorpus\xspace}
\newcommand{\hsplit}{HSplit\xspace}

\graphicspath{ {figures/} }

\title{{ASSET}: {A} Dataset for Tuning and Evaluation of Sentence Simplification Models with Multiple Rewriting Transformations}

\author{Fernando Alva-Manchego$^{1}$\Thanks{ Equal Contribution} \and Louis Martin$^{2,3}$\footnotemark[1] \and Antoine Bordes$^3$ \\ {\bf Carolina Scarton$^1$ \and Benoît Sagot$^2$ \and Lucia Specia$^{1,4}$}\\
$^1$University of Sheffield, $^2$Inria, $^3$Facebook AI Research, $^4$Imperial College London\\
         \texttt{f.alva@sheffield.ac.uk, louismartin@fb.com, abordes@fb.com}\\
         \texttt{c.scarton@sheffield.ac.uk, benoit.sagot@inria.fr}\\
         \texttt{l.specia@imperial.ac.uk}\\
         }

\date{}

\begin{document}
\maketitle
\begin{abstract}
In order to simplify a sentence, human editors perform multiple rewriting transformations: they split it into several shorter sentences,  paraphrase words (i.e.\ replacing complex words or phrases by simpler synonyms), reorder components, and/or delete information deemed unnecessary.
Despite these varied range of possible text alterations, current models for automatic sentence simplification are evaluated using datasets that are focused on a single transformation, such as lexical paraphrasing or splitting.
This makes it impossible to understand the ability of simplification models in more  realistic settings.
To alleviate this limitation, this paper introduces \asset, a new dataset for assessing sentence simplification in English.
\asset is a crowdsourced multi-reference corpus where each simplification was produced by executing several rewriting transformations.
Through quantitative and qualitative experiments, we show that simplifications in \asset are better at capturing characteristics of simplicity when compared to other standard evaluation datasets for the task.
Furthermore, we motivate the need for developing better methods for automatic evaluation using \asset, since we show that current popular metrics may not be suitable when multiple simplification transformations are performed.
\end{abstract}

\section{Introduction}
\label{sec:introduction}
Sentence Simplification (SS) consists in modifying the content and structure of a sentence to make it easier to understand, while retaining its main idea and most of its original meaning~\citep{alvamanchego-etal:2020:survey}. 
Simplified texts can benefit non-native speakers~\cite{paetzold:2016:PHDTHESIS}, people suffering from aphasia~\cite{carroll-etal:1998}, dyslexia~\cite{rello-etal:2013:W4A} or autism~\cite{evans-etal:2014}. 
They also help language processing tasks, such as parsing~\cite{chandrasekar-etal:1996}, summarisation~\cite{silveira-branco:2012}, and machine translation~\cite{hasler-etal:2017}.

In order simplify a sentence, several rewriting transformations can be performed: replacing complex words/phrases with simpler synonyms (i.e.\ lexical paraphrasing), changing the syntactic structure of the sentence (e.g.\ splitting), or removing superfluous information that make the sentence more complicated \cite{petersen:2007,aluisio-etal:2008,bott-saggion:2011:study}.
However, models for automatic SS are evaluated on datasets whose simplifications are not representative of this variety of transformations.
For instance, \turk \cite{xu-etal:2016}, a standard dataset for assessment in SS, contains simplifications produced mostly by lexical paraphrasing, while reference simplifications in HSplit~\cite{sulem-etal:2018:hsplit} focus on splitting sentences.
The Newsela corpus \cite{xu-etal:2015} contains simplifications produced by professionals applying multiple rewriting transformations, but sentence alignments are automatically computed and thus imperfect, and its data can only be accessed after signing a restrictive public-sharing licence and cannot be redistributed, hampering reproducibility.

These limitations in evaluation data prevent studying models' capabilities to perform a broad range of simplification transformations.
Even though most SS models are trained on simplification instances displaying several text transformations (e.g.\ WikiLarge~\cite{zhang-lapata:2017}), we currently do not measure their performance in more \emph{abstractive} scenarios, i.e. cases with substantial modifications to the original sentences.  

In this paper we introduce \textbf{\asset} (\textbf{A}bstractive \textbf{S}entence \textbf{S}implification \textbf{E}valuation and \textbf{T}uning), a new  dataset for tuning and evaluation of automatic SS models. ASSET consists of 23,590 human simplifications associated with the 2,359 original sentences from \turk (10 simplifications per original sentence).
Simplifications in \asset were collected via crowdsourcing (\S~\ref{sec:dataset}), and encompass a variety of rewriting transformations (\S~\ref{sec:transformations}), which make them simpler than those in \turk and \hsplit (\S~\ref{sec:preference}), thus providing an additional suitable benchmark for comparing and evaluating automatic SS models.
In addition, we study the applicability of standard metrics for evaluating SS using simplifications in \asset as references (\S~\ref{sec:evaluating_metrics}). 
We analyse whether BLEU~\cite{papineni-etal:2002:Bleu} or SARI~\cite{xu-etal:2016} scores correlate with human judgements of fluency, adequacy and simplicity, and find that neither of the metrics shows a strong correlation with simplicity ratings. 
This motivates the need for developing better metrics for assessing  SS when multiple rewriting transformations are performed.

We make the following contributions:

\begin{itemize}
    \item A high quality large dataset for tuning and evaluation of SS models containing simplifications produced by applying multiple rewriting transformations.\footnote{\asset is released with a CC-BY-NC license at \\\url{https://github.com/facebookresearch/asset}.}
    \item An analysis of the characteristics of the dataset that turn it into a new suitable benchmark for evaluation.
    \item A study questioning the suitability of popular metrics for evaluating automatic simplifications in a multiple-transformation scenario.
\end{itemize}

\section{Related Work}
\label{sec:related_work}

\subsection{Studies on Human Simplification}

A few corpus studies have been carried out to analyse how humans simplify sentences, and to attempt to determine the rewriting transformations that are performed.

\newcite{petersen-ostendorf:2007} analysed a corpus of 104 original and professionally simplified news articles in English.
Sentences were manually aligned and each simplification instance was categorised as dropped (1-to-0 alignment), split (1-to-N), total (1-to-1) or merged (2-to-1). 
Some splits were further sub-categorised as edited (i.e. the sentence was split and some part was dropped) or different (i.e. same information but very different wording).
This provides evidence that sentence splitting and deletion of information can be performed simultaneously.

\newcite{aluisio-etal:2008} studied six corpora of simple texts (different genres) and a corpus of complex news texts in Brazilian Portuguese, to produce a manual for Portuguese text simplification \cite{specia-etal:2008}. 
It contains several rules to perform the task focused on syntactic alterations: to split adverbial/coordinated/subordinated sentences, to reorder clauses to a subject-verb-object structure, to transform passive to active voice, among others.

\newcite{bott-saggion:2011:study} worked with a dataset of 200 news articles in Spanish with their corresponding manual simplifications. 
After automatically aligning the sentences, the authors determined the simplification transformations performed: change (e.g. difficult words, pronouns, voice of verb), delete (words, phrases or clauses), insert (word or phrases), split (relative clauses, coordination, etc.), proximisation (add locative phrases, change from third to second person), reorder, select, and join (sentences). 

From all these studies, it can be argued that the scope of rewriting transformations involved in the simplification process goes beyond only replacing words with simpler synonyms.
In fact, human perception of complexity is most affected by syntactic features related to sentence structure \cite{brunato-etal:2018:isthissentence}.
Therefore, since human editors make several changes to both the lexical content and syntactic structure of sentences when simplifying them, we should expect that models for automatic sentence simplification can also make such changes.

\subsection{Evaluation Data for SS}

Most datasets for SS \cite{zhu-etal:2010,coster-kauchak:2011,hwang-etal:2015} consist of automatic sentence alignments between related articles in English Wikipedia (EW) and Simple English Wikipedia (SEW).
In SEW, contributors are asked to write texts using simpler language, such as by shortening sentences or by using words from Basic English \cite{ogden:1930}.
However, \newcite{yasseri-etal:2012} found that the syntactic complexity of sentences in SEW is almost the same as in EW.
In addition, \newcite{xu-etal:2015} determined that automatically-aligned simple sentences are sometimes just as complex as their original counterparts, with only a few words replaced or dropped and the rest of the sentences left unchanged. 

More diverse simplifications are available in the Newsela corpus \cite{xu-etal:2015}, a dataset of 1,130 news articles that were each manually simplified to up to 5 levels of simplicity. 
The parallel articles can be automatically aligned at the sentence level to train and test simplification models \cite{alvamanchego-etal:2017:ijcnlp,stajner-etal:2018:CATS}.
However, the Newsela corpus can only be accessed after signing a restrictive license that prevents publicly sharing  train/test splits of the dataset, which impedes reproducibility.

Evaluating models on automatically-aligned sentences is problematic. 
Even more so if only one (potentially noisy) reference simplification for each original sentence is available.
With this concern in mind, \newcite{xu-etal:2016} collected the \turk, a dataset with 2,359 original sentences from EW, each with 8 manual reference simplifications.
The dataset is divided into two subsets: 2,000 sentences for validation and 359 for testing of sentence simplification models.
\turk is suitable for automatic evaluation that involves metrics requiring multiple references, such as BLEU \cite{papineni-etal:2002:Bleu} and SARI \cite{xu-etal:2016}. 
However, \newcite{xu-etal:2016} focused on simplifications through lexical paraphrasing, instructing annotators to rewrite sentences by reducing the number of difficult words or idioms, but without deleting content or splitting the sentences.
This prevents evaluating a model's ability to perform a more diverse set of rewriting transformations when simplifying sentences.
HSplit~\cite{sulem-etal:2018:hsplit}, on the other hand, provides simplifications involving only splitting for sentences in the test set of \turk. 
We build on \turk and HSplit by collecting a dataset that provides several manually-produced simplifications involving multiple types of rewriting transformations.

\subsection{Crowdsourcing Manual Simplifications}
A few projects have been carried out to collect manual simplifications through crowdsourcing. 
\newcite{pellow-eskenazi:2014} built a corpus of everyday documents (e.g.\ driving test preparation materials), and analysed the feasibly of crowdsourcing their sentence-level simplifications.
Of all the quality control measures taken, the most successful was providing a training session to workers, since it allowed to block spammers and those without the skills to perform the task.
Additionally, they proposed to use workers' self-reported confidence scores to flag submissions that could be discarded or reviewed.
Later on, \newcite{pellow-eskenazi:2014:AAAI} presented a preliminary study on producing simplifications through a collaborative process.
Groups of four workers were assigned one sentence to simplify, and they had to discuss and agree on the process to perform it.
Unfortunately, the data collected in these studies is no longer publicly available.

Simplifications in \turk were also collected through crowdsourcing.
Regarding the methodology followed, \newcite{xu-etal:2016} only report  removing bad workers after manual check of their first several submissions.
More recently, \newcite{scarton-etal:2018:SIMPA} used volunteers to collect simplifications for SimPA, a dataset with sentences from the Public Administration domain. 
One particular characteristic of the methodology followed is that lexical and syntactic simplifications were performed independently.

\section{Creating \asset}
\label{sec:dataset}

\begin{table*}[t]
\centering \small
\begin{tabular}{@{}lp{0.87\textwidth}@{}}
\toprule
{\bf Original}    & Their eyes are quite small, and their visual acuity is poor.\\
{\bf \turk}       & Their eyes are very little, and their sight is inferior.\\
{\bf HSplit}      & Their eyes are quite small. Their visual acuity is poor as well.\\
{\bf \asset}        & They have small eyes and poor eyesight.\\
\midrule
{\bf Original}    & His next work, Saturday, follows an especially eventful day in the life of a successful neurosurgeon.\\
{\bf \turk}       & His next work at Saturday will be a successful Neurosurgeon.\\
{\bf HSplit}      & His next work was Saturday. It follows an especially eventful day in the life of a successful Neurosurgeon.\\
{\bf \asset}      & "Saturday" records a very eventful day in the life of a successful neurosurgeon.\\
\midrule
{\bf Original}    & He settled in London, devoting himself chiefly to practical teaching.\\
{\bf \turk}       & He rooted in London, devoting himself mainly to practical teaching.\\
{\bf HSplit}      & He settled in London. He devoted himself chiefly to practical teaching.\\
{\bf \asset}      & He lived in London. He was a teacher.\\
\bottomrule
\end{tabular}
\caption{Examples of simplifications collected for \asset together with their corresponding version from \turk and \hsplit for the same original sentences.}
\label{table:examples}
\end{table*}

We extended \turk~\cite{xu-etal:2016} by using the same original sentences, but crowdsourced manual simplifications that encompass a richer set of rewriting transformations.
Since \turk was adopted as the standard dataset for evaluating SS models, several system outputs on this data are already publicly available \cite{zhang-lapata:2017,zhao-etal:2018:dmass-dcss,martin-etal:2020:controllable}.
Therefore, we can now assess the capabilities of these and other systems in scenarios with varying simplification expectations: lexical paraphrasing with \turk, sentence splitting with HSplit, and multiple transformations with \asset.

\subsection{Data Collection Protocol}
\label{sec:annotation_protocol}

Manual simplifications were collected using Amazon Mechanical Turk (AMT).
AMT allows us to publish HITs (Human Intelligence Tasks), which workers can choose to work on, submit an answer, and collect a reward if the work is approved.
This was also the platform used for \turk.

\paragraph{Worker Requirements.}
Participants were workers who: (1) have a HIT approval rate $>= 95\%$; (2) have a number of HITs approved $> 1000$; (3) are residents of the United States of America, the United Kingdom or Canada; and (4) passed the corresponding Qualification Test designed for our task (more details below).
The first two requirements are measured by the AMT platform and ensure that the workers have experience on different tasks and have had most of their work approved by previous requesters. 
The last two requirements are intended to ensure that the workers have a proficient level of English, and are capable of performing the simplification task.

\paragraph{Qualification Test.}
We provided a training session to workers in the form of a Qualification Test (QT). 
Following \newcite{pellow-eskenazi:2014}, we showed them explanations and examples of multiple simplification transformations (see details below).
Each HIT consisted of three sentences to simplify, and all submissions were manually checked to filter out spammers and workers who could not perform the task correctly.
The sentences used in this stage were extracted from the QATS dataset~\cite{stajner-etal:2016:QATS}.
We had 100 workers take the QT, out of which 42 passed the test (42\%) and worked on the task.

\paragraph{Annotation Round.}
Workers who passed the QT had access to this round. 
Similar to \newcite{pellow-eskenazi:2014}, each HIT now consisted of four original sentences that needed to be simplified.
In addition to the simplification of each sentence, workers were asked to submit confidence scores on their simplifications using a 5-point likert scale (1:Very Low, 5:Very High).
We collected 10 simplifications (similar to \newcite{pellow-eskenazi:2014}) for each of the 2,359 original sentences in \turk.

\paragraph{Simplification Instructions.}
For both the QT and the Annotation Round, workers received the same set of instructions about how to simplify a sentence.
We provided examples of lexical paraphrasing (lexical simplification and reordering), sentence splitting, and compression (deleting unimportant information).
We also included an example where all transformations were performed. 
However, we clarified that it was at their discretion to decide which types of rewriting to execute in any given original sentence.\footnote{Full instructions are available in the dataset's repository.}

Table~\ref{table:examples} presents a few examples of simplifications in \asset, together with references from \turk and \hsplit, randomly sampled for the same original sentences.
It can be noticed that annotators in \asset had more freedom to change the structure of the original sentences.

\subsection{Dataset Statistics}

ASSET contains 23,590 human simplifications associated with the 2,359 original sentences from \turk (2,000 from the validation set and 359 from the test set).
Table~\ref{table:asset_stats_compared} presents some general statistics from simplifications in \asset.
We show the same statistics for \turk and \hsplit for comparison.\footnote{\hsplit is composed of two sets of simplifications: one where annotators were asked to split sentences as much as they could, and one where they were asked to split the original sentence only if it made the simplification easier to read and understand. However, we consider \hsplit as a whole because differences between datasets far outweigh differences between these two sets.}

In addition to having more references per original sentence, \asset's simplifications offer more variability, for example containing many more instances of natural sentence splitting than \turk. 
In addition, reference simplifications are shorter on average in ASSET, given that we allowed annotators to delete information that they considered unnecessary. 
In the next section, we further compare these datasets with more detailed text features.

\begin{table}[htb]
\centering \resizebox{\columnwidth}{!}{%
\begin{tabular}{@{}lrrr@{}}
\toprule
                            & \asset   & \turk      & \hsplit \\
\midrule
Original Sentences          & 2,359     & 2,359     & 359 \\
Num. of References          & 10        & 8         & 4   \\
Type of Simp. Instances     &           &           &      \\
\multicolumn{1}{r}{1-to-1}  & 17,245     & 18,499    & 408   \\
\multicolumn{1}{r}{1-to-N}  & 6,345       & 373       & 1,028  \\
Tokens per Reference        & 19.04     & 21.29     & 25.49 \\
\bottomrule
\end{tabular}
}
\caption{General surface statistics for \asset compared with \turk and \hsplit. A simplification instance is an original-simplified sentence pair.} 
\label{table:asset_stats_compared}
\end{table}

\section{Rewriting Transformations in \asset}
\label{sec:transformations}
We study the simplifications collected for \asset through a series of text features to measure the \emph{abstractiveness} of the rewriting transformations performed by the annotators.
From here on, the analysis and statistics reported refer to the test set only (i.e. 359 original sentences), so that we can fairly compare \asset, \turk and \hsplit.

\begin{figure*}
    \centering
    \includegraphics[width=\linewidth]{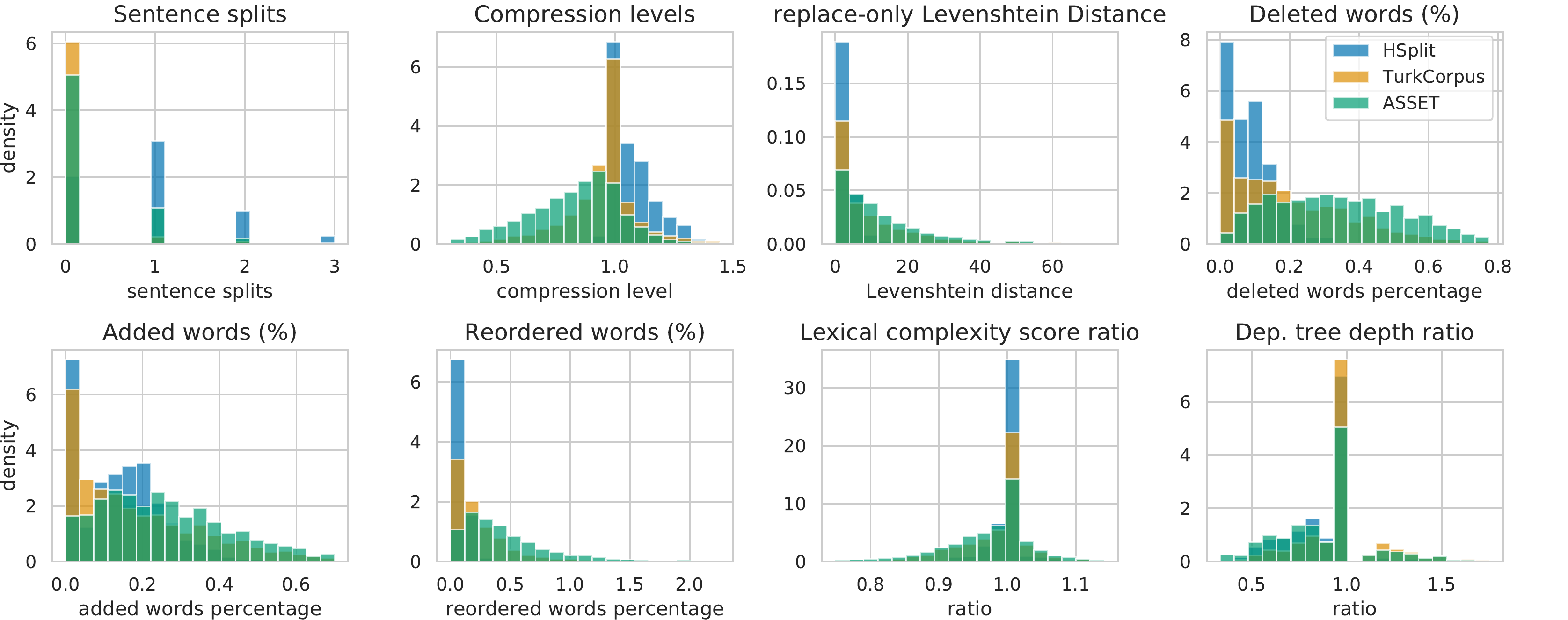}
    \caption{Density of text features in simplifications from \hsplit, \turk, and \asset.}
    \label{figure:dataset_features}
\end{figure*}

\subsection{Text Features}
In order to quantify the rewriting transformations, we computed several low-level features for all simplification instances using the \texttt{tseval} package \cite{qa4ts:martin-etal:2018}:

\begin{itemize}
    \item \textbf{Number of sentence splits:} Corresponds to the difference between the number of sentences in the simplification and the number of sentences in the original sentence. 
    In \texttt{tseval}, the number of sentences is calculated using NLTK \cite{loper2002nltk}. 
    
    \item \textbf{Compression level:} Number of characters in the simplification divided by the number of characters in the original sentence.
    
    \item \textbf{Replace-only Levenshtein distance:} Computed as the normalised character-level Levenshtein distance \cite{levenshtein:1966} for replace operations only, between the original sentence and the simplification.
    Replace-only Levenshtein distance is computed as follows (with $o$ the original sentence and $s$ the simplification): 
    $$\frac{replace\_ops(o, s)}{min(len(o), len(s))}$$
    We do not consider insertions and deletions in the Levenshtein distance computation so that this feature is independent from the compression level.
    It therefore serves as a proxy for measuring the lexical paraphrases of the simplification.
    
    \item \textbf{Proportion of words deleted, added and reordered:} Number of words deleted/reordered from the original sentence divided by the number of words in the original sentence; and the number of words  that were added to the original sentence divided by the number of words in the simplification.
    
    \item \textbf{Exact match:} Boolean feature that equals to true when the original sentence and the simplification are exactly the same, to account for unchanged sentences.
    
    \item \textbf{Word deletion only:} Boolean feature that equals to true when the simplification is obtained only by deleting words from the original sentence.
    This feature captures extractive compression.

    \item \textbf{Lexical complexity score ratio:} We compute the score as the mean squared log-ranks of content words in a sentence (i.e. without stopwords).
    We use the 50k most frequent words of the FastText word embeddings vocabulary \cite{bojanowski2016enriching}. This vocabulary was originally sorted with frequencies of words in the Common Crawl.
    This score is a proxy to the lexical complexity of the sentence given that word ranks (in a frequency table) have been shown to be best indicators of word complexity  \cite{paetzold-specia:2016:semeval}.
    The ratio is then the value of this score on the simplification divided by that of the original sentence. 
    
    \item \textbf{Dependency tree depth ratio:} We compute the ratio of the depth of the dependency parse tree of the simplification relative to that of the original sentence.
    When a simplification is composed by more than one sentence, we choose the maximum depth of all dependency trees.
    Parsing is performed using spaCy.\footnote{\url{github.com/explosion/spaCy}}
    This feature serves as a proxy to measure improvements in structural simplicity.
\end{itemize}

Each feature was computed for all simplification instances in the dataset and then aggregated as a histogram (Figure~\ref{figure:dataset_features}) and as a percentage (Table~\ref{table:abscorpus_vs_turkcorpus}).

\subsection{Results and Analysis}

\begin{table}[tb]
\centering \resizebox{\columnwidth}{!}{%
\begin{tabular}{@{}lrrr@{}}
\toprule
                       & \asset     & \turk      & \hsplit \\
\midrule
Sentence Splitting     & 20.2\%     & 4.6\%      & 68.2\%       \\
Compression ($<$75\%)  & 31.2\%     & 9.9\%      & 0.1\%       \\
Word Reordering        & 28.3\%    & 19.4\%     & 10.1\%     \\
Exact Match            & 0.4\%      & 16.3\%     & 26.5\%     \\
Word Deletion Only     & 4.5\%      & 3.9\%      & 0.0\%     \\
\bottomrule
\end{tabular}
}
\caption{Percentage of simplifications featuring one of different rewriting transformations operated in \asset, \turk and \hsplit. A simplification is considered as compressed when its character length is less than 75\% of that of the original sentence.} 
\label{table:abscorpus_vs_turkcorpus}
\end{table}

Figure~\ref{figure:dataset_features} shows the density of all features in \asset, and compares them with those in \turk and \hsplit.
Table~\ref{table:abscorpus_vs_turkcorpus} highlights some of these statistics.
In particular, we report the percentage of sentences that: have at least one sentence split, have a compression level of 75\% or lower, have at least one reordered word, are exact copies of the original sentences, and operated word deletion only (e.g. by removing only an adverb).

Sentence splits are practically non-existent in \turk (only 4.6\% have one split or more), and are more present and distributed in \hsplit.
In \asset, annotators tended to not split sentences, and those who did mostly divided the original sentence into just two sentences (1 split).

Compression is a differentiating feature of \asset. 
Both \turk and \hsplit have high density of a compression ratio of 1.0, which means that no compression was performed. 
In fact, \hsplit has several instances with compression levels greater than 1.0, which could be explained by splitting requiring adding words to preserve fluency.
In contrast, \asset offers more variability, perhaps signalling that annotators consider deleting information as an important simplification operation.

By analysing replace-only Levenshtein distance, we can see that simplifications in \asset paraphrase the input more. 
For \turk and \hsplit, most simplifications are similar to their original counterparts (higher densities closer to 0).
On the other hand, \asset's simplifications are distributed in all levels, indicating more diversity in the rewordings performed.
This observation is complemented by the distributions of deleted, added and reordered words.
Both \turk and \hsplit have high densities of ratios close to 0.0 in all these features, while \asset's are more distributed.
Moreover, these ratios are rarely equal to 0 (low density), meaning that for most simplifications, at least some effort was put into rewriting the original sentence. This is comfirmed by the low percentage of exact matches in \asset (0.4\%) with respect to \turk (16.3\%) and \hsplit (26.5\%).
Once again, it suggests that more rewriting transformations are being performed in \asset. 

In terms of lexical complexity, \hsplit has a high density of ratios close to 1.0 due to its simplifications being structural and not lexical.
\turk offers more variability, as expected, but still their simplifications contain a high number of words that are equally complex, perhaps due to most simplifications just changing a few words. 
On the other hand, \asset's simplifications are more distributed across different levels of reductions in lexical complexity.

Finally, all datasets show high densities of a 1.0 ratio in dependency tree depth.
This could mean that significant structural changes were not made, which is indicated by most instances corresponding to operations other than splitting. 
However, \asset still contains more simplifications that reduce syntactic complexity than \turk and \hsplit.

\section{Rating Simplifications in \asset}
\label{sec:preference}

Here we measure the quality of the collected simplifications using human judges.
In particular, we study if the \emph{abstractive} simplifications in \asset (test set) are preferred over lexical-paraphrase-only or splitting-only simplifications in \turk (test set) and \hsplit, respectively.

\subsection{Collecting Human Preferences}
\label{sec:qualification_test_rating}
Preference judgments were crowdsourced with a protocol similar to that of the simplifications (\S~\ref{sec:annotation_protocol}). 

\paragraph{Selecting Human Judges.}
Workers needed to comply with the same basic requirements as described in \S~\ref{sec:annotation_protocol}.
For this task, the Qualification Test (QT) consisted in rating the quality of simplifications based on three criteria: fluency (or grammaticality), adequacy (or meaning preservation), and simplicity.
Each HIT consisted of six original-simplified sentence pairs, and workers were asked to use a continuous scale (0-100) to submit their level of agreement (0: Strongly disagree, 100: Strongly agree) with the following statements:

\begin{enumerate}
    \item The Simplified sentence adequately expresses the meaning of the Original, perhaps omitting the least important information.
    \item The Simplified sentence is fluent, there are no grammatical errors.
    \item The Simplified sentence is easier to understand than the Original sentence.
\end{enumerate}

Using continuous scales when crowdsourcing human evaluations is common practice in Machine Translation \cite{bojar-etal:2018:wmt18,barrault-etal:2019:wmt19}, since it results in higher levels of inter-annotator consistency \cite{graham-etal-2013:continuous}.
The six sentence pairs for the Rating QT consisted of: 

\begin{itemize}
    \item Three submissions to the Annotation QT, manually selected so that one contains splitting, one has a medium level of compression, and one contains grammatical and spelling mistakes.
    These allowed to check that the particular characteristics of each sentence pair affect the corresponding evaluation criteria.
    \item One sentence pair extracted from WikiLarge~\cite{zhang-lapata:2017} that contains several sentence splits.
    This instance appeared twice in the HIT and allowed checking for intra-annotator consistency.
    \item One sentence pair from WikiLarge where the Original and the Simplification had no relation to each other. 
    This served to check the attention level of the worker.  
\end{itemize}

All submitted ratings were manually reviewed to validate the quality control established and to select the qualified workers for the task.

\paragraph{Preference Task.}
For each of the 359 original sentences in the test set, we randomly sampled one reference simplification from \asset and one from \turk, and then asked qualified workers to choose which simplification answers best each of the following questions:

\begin{itemize}
    \item \textbf{Fluency}: Which sentence is more fluent?
    \item \textbf{Meaning}: Which sentence expresses the original meaning the best?
    \item \textbf{Simplicity}: Which sentence is easier to read and understand?
\end{itemize}

Workers were also allowed to judge simplifications as ``similar'' when they could not determine which one was better.
The same process was followed to compare simplifications in \asset  against those in \hsplit.
Each HIT consisted of 10 sentence pairs. %

\subsection{Results and Analysis}

Table~\ref{table:asset_vs_others_human} (top section) presents, for each evaluation dimension, the percentage of times a simplification from \asset or \turk was preferred over the other, and the percentage of times they were judged as ``similar''.
In general, judges preferred \asset's simplifications in terms of fluency and simplicity.
However, they found \turk' simplifications more meaning preserving.
This is expected since they were produced mainly by replacing words/phrases with virtually no deletion of content.

A similar behaviour was observed when comparing \asset to \hsplit (bottom section of Table~\ref{table:asset_vs_others_human}).
In this case, however, the differences in preferences are greater than with \turk.
This could indicate that changes in syntactic structure are not enough for a sentence to be consider simpler.

\begin{table}[tb]
\centering \resizebox{0.84\columnwidth}{!}{%
\begin{tabular}{@{}lrrr@{}}
\toprule
 &  Fluency &  Meaning & Simplicity \\
\midrule
\asset & \textbf{38.4\%*}  & 23.7\% & \textbf{41.2\%*} \\
\turk & 22.8\% & \textbf{37.9\%*} & 20.1\% \\
Similar & 38.7\% & 38.4\% & 38.7\% \\
\midrule
\asset  & \textbf{53.5\%}*   & 17.0\%          & \textbf{59.0\%}* \\
\hsplit & 19.5\%            & \textbf{51.5\%}* & 14.8\% \\
Similar & 27.0\%            & 31.5\%          & 26.2\% \\
\bottomrule
\end{tabular}
}
\caption{Percentages of human judges who preferred simplifications in \asset or \turk, and \asset or \hsplit, out of 359 comparisons. * indicates a statistically significant difference between the two datasets (binomial test with p-value $<0.001$).}
\label{table:asset_vs_others_human}
\end{table}

\section{Evaluating Evaluation Metrics}
\label{sec:evaluating_metrics}

In this section we study the behaviour of evaluation metrics for SS when using \asset's simplifications (test set) as references.
In particular, we measure the correlation of standard metrics with human judgements of fluency, adequacy and simplicity, on simplifications produced by automatic systems. 

\subsection{Experimental Setup}

\paragraph{Evaluation Metrics.}
We analysed the behaviour of two standard metrics in automatic evaluation of SS outputs: BLEU~\cite{papineni-etal:2002:Bleu} and SARI \cite{xu-etal:2016}. 
BLEU is a precision-oriented metric that relies on the number of $n$-grams in the output that match $n$-grams in the references, independently of position.
SARI measures improvement in the simplicity of a sentence based on the $n$-grams added, deleted and kept by the simplification system. 
It does so by comparing the output of the simplification model to multiple references and the original sentence, using both precision and recall. 
BLEU has shown positive correlation with human judgements of grammaticality and meaning preservation \cite{stajner-EtAl:2014:PITR,wubben-etal:2012,xu-etal:2016}, while SARI has high correlation with judgements of simplicity gain \cite{xu-etal:2016}.
In our experiments, we used the implementations of these metrics available in the EASSE package for automatic sentence simplification evaluation \cite{alvamanchego-etal:2019:easse}.\footnote{\url{https://github.com/feralvam/easse}}
We computed all the scores at sentence-level as in the experiment by \newcite{xu-etal:2016}, where they compared sentence-level correlations of FKGL, BLEU and SARI with human ratings. 
We used a smoothed sentence-level version of  BLEU so that comparison is possible, even though BLEU was designed as a corpus-level metric.

\paragraph{System Outputs.}
We used publicly-available simplifications produced by automatic SS systems: PBSMT-R~\cite{wubben-etal:2012}, which is a phrase-based MT model; Hybrid~\cite{narayan-gardent:2014}, which uses phrase-based MT coupled with semantic analysis; SBSMT-SARI~\cite{xu-etal:2016}, which relies on syntax-based MT; NTS-SARI~\cite{nisioi-EtAl:2017:NTS}, a neural sequence-to-sequence model with a standard encoder-decoder architecture; and ACCESS \cite{martin-etal:2020:controllable}, an encoder-decoder architecture conditioned on explicit attributes of sentence simplification.

\paragraph{Collection of Human Ratings.}
We randomly chose 100 original sentences from \asset and, for each of them, we sampled one system simplification.
The automatic simplifications were selected so that the distribution of simplification transformations (e.g.\ sentence splitting, compression, paraphrases) would match that from human simplifications in \asset.
That was done so that we could obtain a sample that has variability in the types of rewritings performed.
For each sentence pair (original and automatic simplification), we crowdsourced 15 human ratings on fluency (i.e. grammaticality), adequacy (i.e. meaning preservation) and simplicity, using the same worker selection criteria and HIT design of the Qualification Test as in \S~\ref{sec:qualification_test_rating}.

\subsection{Inter-Annotator Agreement}
We followed the process suggested in \cite{graham-etal-2013:continuous}.
First, we normalised the scores of each rater by their individual mean and standard deviation, which helps eliminate individual judge preferences.
Then, the normalised continuous scores were converted to five interval categories using equally spaced bins.
After that, we followed \newcite{pavlick-tetreault:2016} and computed quadratic weighted Cohen's $\kappa$~\cite{cohen:1968:kappa-weighted} simulating two raters: for each sentence, we chose one worker's rating as the category for annotator A, and selected the rounded average scores for the remaining workers as the category for annotator B. 
We then computed $\kappa$ for this pair over the whole dataset. 
We repeated the process 1,000 times to compute the mean and variance of $\kappa$.
The resulting values are: $0.687 \pm 0.028$ for Fluency, $0.686 \pm 0.030$ for Meaning and $0.628 \pm 0.032$ for Simplicity.
All values point to a moderate level of agreement, which is in line with the subjective nature of the simplification task.

\subsection{Correlation with Evaluation Metrics}

\begin{table}[tb]
\centering \resizebox{\columnwidth}{!}{%
\begin{tabular}{@{}llrrr@{}}
\toprule
Metric  & References &      Fluency  &  Meaning & Simplicity \\
\midrule
BLEU    & \asset  & 0.42* &   0.61* &      0.31* \\
        & \turk &  0.35* &   0.59* &       0.18 \\
SARI    & \asset  &  0.16 &    0.13 &      0.28* \\
        & \turk &   0.14 &    0.10 &       0.17 \\
\bottomrule
\end{tabular}
}
\caption{Pearson correlation of human ratings with \textbf{automatic metrics} on system simplifications. * indicates a significance level of p-value $<0.05$.\label{table:correlation_metric}}
\end{table}

We computed the Pearson correlation between the normalised ratings and the evaluation metrics of our interest (BLEU and SARI) using \asset or \turk as the set of references.
We refrained from experimenting with \hsplit since neither BLEU nor SARI correlate with human judgements when calculated using that dataset as references \cite{sulem-etal:2018:hsplit}.
Results are reported in Table~\ref{table:correlation_metric}.

BLEU shows a strong positive correlation with Meaning Preservation using either simplifications from \asset or \turk as references. 
There is also some positive correlation with Fluency judgements, but that is not always the case for Simplicity: no correlation when using \turk and moderate when using \asset.
This is in line with previous studies that have shown that BLEU is not a good estimate for simplicity \cite{wubben-etal:2012,xu-etal:2016,sulem-etal:2018}.

In the case of SARI, correlations are positive but low with all criteria and significant only for simplicity with \asset's references.
\newcite{xu-etal:2016} showed that SARI correlated with human judgements of simplicity gain, when instructing judges to \textit{``grade the quality of the variations by identifying the words/phrases that are altered, and counting how many of them are good simplifications''}.\footnote{ \url{https://github.com/cocoxu/simplification/tree/master/HIT_MTurk_crowdsourcing}} 
The judgements they requested differ from the ones we collected, since theirs were tailored to rate simplifications produced by lexical paraphrasing only.
These results show that SARI might not be suitable for the evaluation of automatic simplifications with multiple rewrite operations. %

In Table~\ref{table:correlation_feauture}, we further analyse the human ratings collected, and compute their correlations with similar text features as in \S~\ref{sec:transformations}.
The results shown reinforce our previous observations that judgements on Meaning correlate with making few changes to the sentence: strong negative correlation with Levenshtein distance, and strong negative correlation with proportion of words added, deleted, and reordered.
No conclusions could be drawn with respect to Simplicity.

\begin{table}[tb]
\centering \resizebox{\columnwidth}{!}{%
\begin{tabular}{@{}lrrr@{}}
\toprule
Feature                     &  Fluency  &  Meaning & Simplicity \\
\midrule
Length                      &    0.12 &   0.31* &       0.03 \\
Sentence Splits             &   -0.13 &   -0.06 &      -0.08 \\
Compression Level           &   0.26* &   0.46* &       0.04 \\
Levenshtein Distance        &   -0.40*&  -0.67* &      -0.18 \\
Replace-only Lev. Dist.     &   -0.04 &	 -0.17  &      -0.06 \\
Prop. Deleted Words         &  -0.43* &  -0.67* &      -0.19 \\
Prop. Added Words           &   -0.19 &  -0.38* &      -0.12 \\
Prop. Reordered Words       &  -0.37* &  -0.57* &      -0.18 \\
Dep. Tree Depth Ratio       &    0.20 &    0.24 &       0.06 \\
Word Rank Ratio             &    0.04 &    0.08 &      -0.05 \\
\bottomrule
\end{tabular}
}
\caption{Pearson correlation of human ratings with \textbf{text features} on system simplifications. * indicates a significance level of p-value $<0.01$.\label{table:correlation_feauture}}
\end{table}

\section{Conclusion}
\label{sec:conclusion}

We have introduced \asset, a new dataset for tuning and evaluation of SS models.
Simplifications in \asset were crowdsourced, and annotators were instructed to apply multiple rewriting transformations. 
This improves current publicly-available evaluation datasets, which are focused on only one type of transformation.
Through several experiments, we have shown that \asset contains simplifications that are more \emph{abstractive}, and that are consider simpler than those in other evaluation corpora.
Furthermore, we have motivated the need to develop new metrics for automatic evaluation of SS models, especially when evaluating simplifications with multiple rewriting operations.
Finally, we hope that ASSET's multi-transformation features will motivate the development of SS models that benefit a variety of target audiences according to their specific needs such as people with low literacy or cognitive disabilities.

\section*{Acknowledgements}

This work was partly supported by Benoît Sagot's chair in the PRAIRIE institute, funded by the French national agency ANR as part of the ``Investissements d’avenir'' programme under the reference \mbox{ANR-19-P3IA-0001}.

\bibliographystyle{acl_natbib}
\bibliography{bibliography}

\end{document}